\documentclass{article}
\usepackage{amsmath}
\usepackage{makecell}

 \usepackage[sglblindworkshop, final]{neurips_2025}
\workshoptitle{Machine Learning for the Physical Sciences}

\usepackage[utf8]{inputenc} 
\usepackage[T1]{fontenc}    
\usepackage{hyperref}       
\usepackage{url}            
\usepackage{booktabs}       
\usepackage{amsfonts}       
\usepackage{nicefrac}       
\usepackage{microtype}      
\usepackage{xcolor}         
\usepackage{graphicx}

\usepackage{caption}
\usepackage{float}

\title{IonCast: A Deep Learning Framework for Forecasting Ionospheric Dynamics}

\author{%
Halil S.~Kelebek \\
  Department of Engineering Science \\
  University of Oxford, UK \\
  \texttt{halil@robots.ox.ac.uk} \\
  \And
  Linnea M.~Wolniewicz \\
  Department of Information and Computer Science \\
  University of Hawai`i at Mānoa, USA \\
  \texttt{linneamw@hawaii.edu} \\
  \And
  Michael D.~Vergalla \\
  Free Flight Research Lab \\
  Sunnyvale, USA \\
  \texttt{mike@freeflightlab.org} \\
  \And
  Simone Mestici \\
  Università degli Studi di Roma Sapienza \\
  Rome, Italy \\
  \texttt{simone.mestici@uniroma1.it} \\
  \And
  Giacomo Acciarini \\
  European Space Agency (ESA) \\
  \texttt{giacomo.acciarini@esa.int} \\
  \And
  Bala Poduval \\
  University of New Hampshire \\
  \And
  Olga Verkhoglyadova \\
  NASA Jet Propulsion Laboratory \\
  \And
  Madhulika Guhathakurta \\
  NASA Headquarters \\
  \texttt{madhulika.guhathakurta@nasa.gov}
  \And
  Thomas E. Berger \\
  Space Weather Technology, Research, and Education Center  \\
  University of Colorado Boulder \\
  \texttt{Thomas.Berger@colorado.edu} \\
  \And
  Frank Soboczenski \\
  Department of Computer Science \\
  University of York \& King’s College London, UK \\
  \texttt{frank.soboczenski@york.ac.uk} \\
  \And
  Atılım Güneş Baydin \\
  Department of Computer Science \\
  University of Oxford, UK \\
  \texttt{gunes@robots.ox.ac.uk} \\
}

\begin{document}

\maketitle 


\begin{abstract} 
The ionosphere is a critical component of near-Earth space, shaping GNSS accuracy, high-frequency communications, and aviation operations. For these reasons, accurate forecasting and modeling of ionospheric variability has become increasingly relevant. To address this gap, we present IonCast, a suite of deep learning models that include a GraphCast-inspired model tailored for ionospheric dynamics. IonCast leverages spatiotemporal learning to forecast global Total Electron Content (TEC), integrating diverse physical drivers and observational datasets. Validating on held-out storm-time and quiet conditions highlights improved skill compared to persistence. By unifying heterogeneous data with scalable graph-based spatiotemporal learning, IonCast demonstrates how machine learning can augment physical understanding of ionospheric variability and advance operational space weather resilience. 

\end{abstract}

\section{Introduction}

The Earth’s ionosphere, extending from $\sim$ 50 to 1500 km altitude, is a partially ionized region of the upper atmosphere shaped by solar radiation, magnetospheric convection, and thermospheric dynamics \cite{kelley2009}. The complex coupling within near-Earth space drives a wide spectrum of phenomena, from small-scale plasma irregularities \cite{Tsunoda1988} to large-scale geomagnetic storms that have tangible consequences for modern society. Ionospheric disturbances, for example, can significantly degrade the accuracy of Global Navigation Satellite Systems (GNSS) \cite{Kintner1976}, while geomagnetic storms generate geomagnetically induced currents that pose serious risks to power grids \cite{Pulkkinen2017} and can also compromise orbital stability in Low Earth Orbit \cite{Kataoka2022}. As modern society reliance on space-based infrastructure grows, accurate ionospheric forecasting has become essential \cite{Berger2020}.

The Total Electron Content (TEC), defined as the integral of the electron density along the ray path between two points, is one of the most important descriptors of the state of the ionosphere \cite{Jakowski2011}. The Jet Propulsion Laboratory (JPL) generates Global Ionospheric Maps (GIM) of TEC with a temporal cadence of 15 minutes and provides these products as part of a near–real-time service \cite{Martire2024}. These resources are fundamental to understand the global spatio-temporal evolution of TEC and serve as a foundation for building data-driven models.\\ Several empirical and physics-based models have been developed to generate local or GIM of TEC as a function of solar and geomagnetic drivers. Empirical models such as the International Reference Ionosphere (IRI) \cite{Bilitza2011,Bilitza2017} and physics-based models such as the Global Ionosphere-Thermosphere Model (GITM; \cite{Ridley2006}) have been instrumental in advancing ionospheric research, but carry intrinsic limitations that have been well documented \cite{Bilitza2022}.

The increasing availability of high-quality ionospheric datasets make research in this field particularly well-suited for machine learning (ML) applications. Although still in its early stages, several studies have demonstrated the promise of ML-based TEC modeling. Current approaches rely on classical methods such as gradient boosting (XGBoost) or Multi-Layer Perceptrons (MLP) \cite{Zhukov2021, Smirnov2023}, while more advanced methods like Bi-directional Long-Term-Short-Term Memory (BiLSTM) \cite{Xiong2021} remain restricted to narrow geographic regions. A recent series of work also investigated the application of deep learning architectures to model and forecast global TEC using several physically relevant inputs \cite{liu2020, ren2023}. These works have shown that deep learning architectures are indeed capable of consistently reproducing the complex ionospheric behavior.

This context highlights the need for advanced ML architectures that can process heterogeneous and multi-source data, operate at global scales, and deliver reliable nowcasting, i.e. the model learns spatiotemporal patterns to project the current state forward in time, and forecasting capabilities. To address this challenge, we present IonCast, a modular framework built on state-of-the-art ML methods. IonCast is designed as a global, data-driven nowcast and forecast model of the TEC, aiming to complement and potentially substitute empirical and physics-based approaches. Model code and data used to produce the IonCast models can be found on GitHub \footnote{https://github.com/FrontierDevelopmentLab/2025-HL-Ionosphere}.

\section{Methods}
\subsection{Data}
\label{sec:data}

For this work a combination of space- and ground-based data was processed, integrated to make it machine learning ready. In particular, the dataset curated consists of \(2\text{D}\) vertical TEC global maps and \(1\text{D}\) driver time series. The \(2\text{D}\) component consists of JPL Global Ionospheric Maps (GIM) of vertical TEC at 15-min cadence (JPL/IGS service; \cite{Mannucci1998, Martire2024, jpld}). The \(1\text{D}\) drivers include solar-wind and geomagnetic parameters (e.g., SYM-H/ASY-D, IMF (\(B_{xyz}\)) , V$_{sw}$) from NASA/GSFC OMNI2 via OMNIWeb (\cite{Stone1998, Nishida1992, Wilson2021}); planetary activity indices (\(K_p, A_p\)) (compiled from NOAA SWPC/GFZ; \cite{Matzka21}); and solar irradiance proxies (F10.7 and S10.7/M10.7/Y10.7) \cite{Tobiska2012}. Auxiliary spatial features use quasi-dipole magnetic coordinates \cite{Laundal2017} and orbital mechanics data (Sun/Moon ephemerides: subsolar/sublunar points, zenith angles, Earth–Sun/Moon distances; \cite{Giorgini1996}). All sources are publicly accessible via the cited archives; processed outputs are released with this work as standardized \(2\text{D}\) maps and aligned \(1\text{D}\) time series. To the best of our knowledge, this dataset is one of the most comprehensive assembled for this type of analysis. This dataset's inclusion of auxiliary spatial features allows us to investigate the impact of providing the model further physical information, which was not attempted in previous ML applications.

\begin{figure}
    \noindent
    \center
    \makebox[\textwidth][c]{\includegraphics[width=1.25\textwidth]{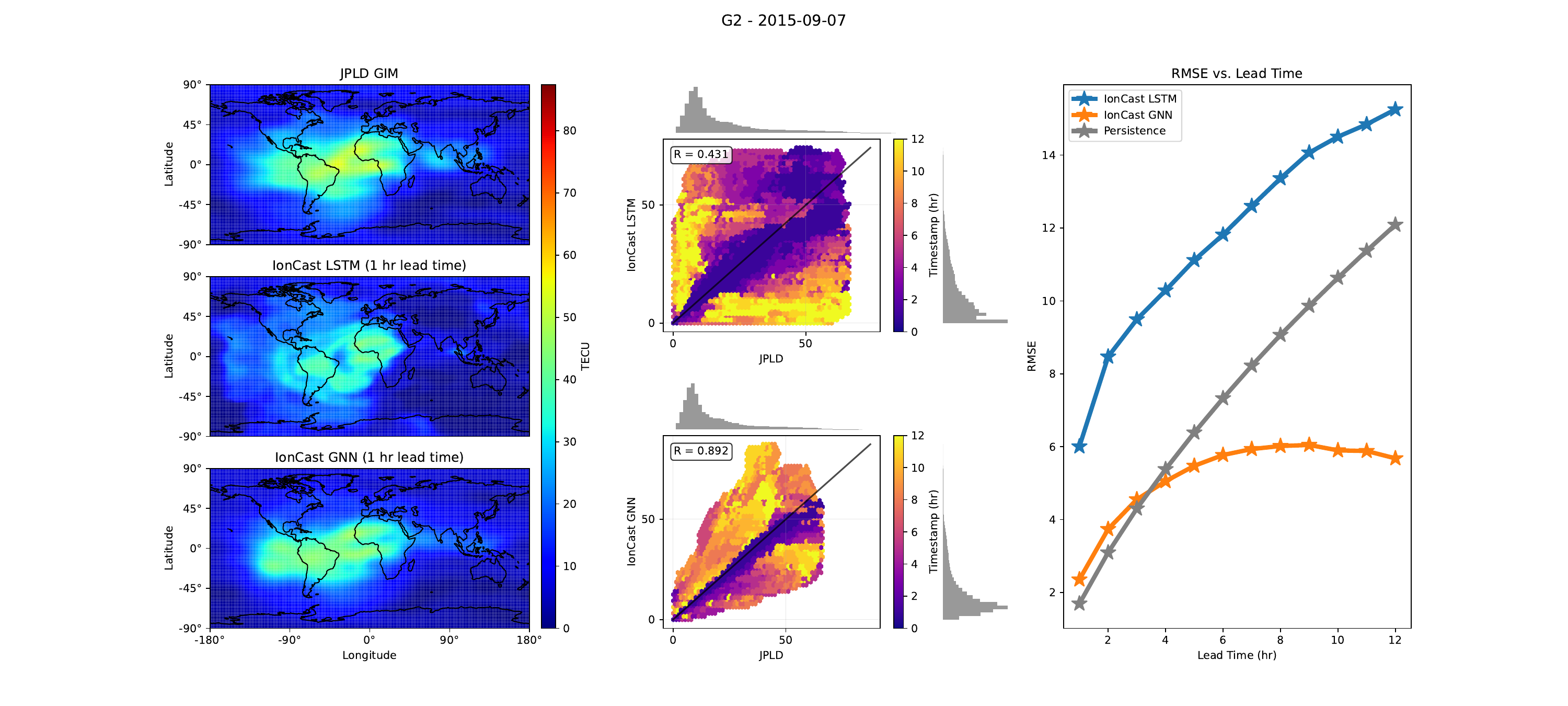}}
    \caption{Model predictions plotted and evaluated over various forecast lead times for a moderate geomagnetic storm (G2 event). \emph{Left:} Subplots show ground truth data from JPL (upper) plotted against predictions by the IonCast LSTM (middle) and IonCast GNN (lower) models at a 60-minute lead time. The colorbar has units TECU (TEC Units). \emph{Center:} The joint distribution between JPL (x-axis) and predictions (y-axis) over a 12-hour long-horizon forecast. The diagonal indicates perfect prediction; points below (above) the diagonal correspond to underprediction (overprediction). The colorbar encodes the mode of the forecast time (minutes) at which the prediction occurs, highlighting the error growth across the sequence. Marginal histograms show the distributions of ground truth and predicted values throughout the forecast. \emph{Right:} The plot compares the average global RMSE of the IonCast LSTM (blue), IonCast GNN (orange), and persistence model predictions across growing forecast horizons. }
    \label{fig:results-moderate}
\end{figure}
\subsection{Models}

We evaluate the forecasting performance of two models for total electron content (TEC): a long short-term memory (LSTM) \citep{Hochreiter1997} baseline, IonCast LSTM, and a Graph-based architecture inspired by GraphCast \citep{Lam2023}, IonCast GNN. Both models autoregressively predict TEC using JPL \cite{Mannucci1998, Martire2024, jpld} data products and learn the dynamics of the ionosphere well. Model hyperparameters were tuned using WandB \cite{wandb} hyperparameter sweeps.

The IonCast LSTM model uses a convolutional encoder–decoder architecture with an LSTM bottleneck, where the convolutional neural network (CNN) encodes each two-dimensional TEC map into a latent representation, the LSTM maintains a hidden state across timesteps, and the CNN decoder reconstructs the forecasted sequence. This model uses a six-layer convolutional LSTM with circular padding to downsample 180x360 ionospheric maps (and other data sources included as extra image channels) to a 128-dimensional embedding and process the temporal sequences of the embeddings. The decoder then reconstructs predictions using bilinear upsampling and transposed convolutions back to the original resolution. IonCast LSTM is trained with a batch size of 4, dropout rate of 0.15, learning rate of 2e-4, and JPLD loss weight of 20.

The IonCast GNN model is based on Google's GraphCast architecture implemented in PyTorch by the NVIDIA PhysicsNeMo framework \citep{physicsnemo}. The architecture follows an encoder, processor, decoder structure. The encoder performs message passing on a graph that converts from the latitude-longitude grid to a spherical icosahedral mesh, the processor learns via message passing on a multi-layer icosahedral mesh, and finally, the decoder learns on the graph mapping from the mesh back to a latitude-longitude grid. IonCast GNN takes as input all the information from previous states and global parameters as grid node features in the initial grid-to-mesh graph. To generate autoregressive forecasts, IonCast GNN predicts all input variables except those identified as forcings. Forcings are features whose values are known or can be analytically computed for any given timestamp, be it in the past or future. In the case of IonCast GNN, the forcings include all orbital mechanics-derived quantities, such as the positions of the Sun and Moon. During both training and inference, the model is provided with the ground truth forcings. These forcing channels are concatenated as additional features to each grid node, including those corresponding to the prediction timesteps. For all non-forcing channels, which include the TEC maps, the 1D solar and geomagnetic drivers, and quasi-dipole magnetic coordinates, IonCast GNN operates autoregressively. The model receives as input the ground truth forcing and non-forcing values for timestamps within the context window, up to time $t$. From the ground truth context, the model autoregressively generates predictions for times $t + 1$ up to $t + k$ with no additional ground truth values for the non-forcing channels. The IonCast GNN model is trained with six multi-mesh levels and six processor layers, with message passing occurring between the 32-hop neighbors of each node. The model is trained with a batch size of 1, dropout rate of 0.15, learning rate of 3e-4, and JPLD loss weight of 2.

Both models are trained on the data detailed in Section \ref{sec:data}, covering the time period from 2010-05-13T00:00:00 to 2024-08-01T00:00:00 at 15-minute cadence, to align with JPL-GIM data. However, to train in a computationally efficient manner, the IonCast LSTM and GNN models train on every 256th sequence of 2 hours (sequences skip 2.66 days between start and end dates). Training is conducted on the Google Cloud Platform. Test and validation data are removed from the training set for specified dates that span various levels of geomagnetic storms, as defined by the NOAA geomagnetic storm scale (G0, G1, G2, G3, G4, and G5), for a total of 10\% of storms at each scale removed from the training set. Models are trained to minimize the mean squared error between predictions and ground truth values, for all targets except forcing features. Both the IonCast LSTM and IonCast GNN models are trained with context windows of 8 (3 hours) to predict 1 timestep (15 minutes) ahead, and predict residual targets ($x_{T+1} = x_T + \hat{x}_{predicted}$).

\section{Results}

Model performance is evaluated by comparing forecasted predictions against the GIM released by JPL \citep{jpld}, which are processed from GNSS stations. To benchmark the forecast performance of our models against JPL TEC maps, we compute the root mean squared error (RMSE) over geomagnetic test events withheld during model training. The RMSE is computed over the full event duration, latitude, and longitude TEC predictions compared with JPL TEC maps. The performance of our models across various forecast lead times for a moderate storm event (G2) is shown in Figure~\ref{fig:results-moderate}. Qualitatively, both models properly reconstruct the main features of the global TEC, including the equatorial enhancement, with the IonCast GNN model reconstructing the global structure of the TEC more accurately than the LSTM.

As a baseline, we use a persistence model, which takes the TEC map from the previous frame as the prediction for the TEC map at a given lead time. The IonCast GNN models outperforms persistence for almost all forecast lead times, with the performance over the persistence improving as the forecast horizon increases, as shown by the orange line in the RMSE plot dipping below the gray. The IonCast LSTM model struggles at longer horizon forecasts compared to the GNN, having a bias towards over-predicting the TEC as the forecast proceeds, whereas the RMSE for IonCast GNN stabilizes from hours 6 to 12. This is visible in the ground truth vs. prediction hexbin plots by the dark colors, which correspond to short lead times, falling along the perfect prediction (diagonal line), and lighter colors corresponding to larger lead times moving further from the diagonal line, especially for the LSTM model. 

Model performance across different latitudinal bands and storm levels is examined in Appendix Figure \ref{fig:results-quiet-moderate-severe}, and a better-performing IonCast LSTM model that beats persistence is shown in Appendix Figure \ref{fig:results-moderate-goodlstm}. Additionally, we compare the performance of our models across growing forecast horizons to the state-of-the-art empirical model, IRI \cite{Bilitza2011}, in Appendix Table \ref{tab:iri_comparison}. We find that the IonCast GNN model outperforms IRI up to a forecast horizon of 6 hours, however, it is not a perfect comparison as IRI is not a forecast model.

To assess the contribution of different data sources to model performance, we performed an ablation study by isolating the contribution of each data source to IonCast GNN. We selected the combination of input features presented in Table~\ref{tab:ablation_inputs} to enhance the understanding of how different classes of physical inputs contribute to forecasting TEC. For these reasons, the classes of solar radiation, solar wind, geomagnetic indices, and orbital mechanics were separated. The models were trained at a 15-minute cadence with a 2-hour context window and evaluated on a 12-hour-long horizon forecast over JPLD predictions. All data streams other than JPLD + F10.7 outperform the model trained solely on a TEC input, suggesting they play a role in aiding the TEC prediction. The poor performance of the F10.7 model may be due to the fact that the F10.7 parameter is produced at a daily cadence, which varies too slowly to provide useful information for forecasts over a 12-hour horizon at a 15-minute temporal resolution. The main contribution to model performance was from the inclusion of the orbital mechanics and quasi-dipole magnetic coordinates, even outperforming the model trained on all data streams. The importance of the orbital mechanics and magnetic coordinate maps is that they provide the model context on the trajectory of the TEC as the Earth rotates, through the encoding of the subsolar and sublunar points as well as the structure of the Earth's magnetic field. The models trained without these channels struggle to account for the apparent motion of the ionosphere over longer forecast horizons (>4 hours), resulting in an eventual spatial drift, explaining the increase in RMSE. One reason the model trained on the JPLD + orbital mechanics + quasi-dipole outperforms the model trained on all datastreams may be caused by the reduced number of channels over which the loss is computed, as the orbital mechanics are forcings and aren't included in the model loss. Thus the model is free to focus on optimizing the TEC channel performance. This may suggest that the performance gained by the inclusion auxiliary channel is not great enough to compensate for the increased complexity in the optimization. Finally, using a residual target was shown to provide a notable improvement in model performance.
\begin{table*}[h!]
\centering
\caption{Ablation study over the influence of input data. RMSE values are averaged over a 12-hour forecast on the validation set. Bold values highlight the experiment with the lowest RMSE value.}

\begin{tabular}{lc}
\hline

\textbf{Input Features} & \textbf{RMSE (TECU)} \\
\hline
JPLD                                   & $22.4 \pm 3.2$ \\
JPLD + F10.7                           & $23.9 \pm 10.2$ \\
JPLD + F10.7, S10.7, M10.7, JB2008     & $13.3 \pm 4.4$\\
JPLD + Ap \& Kp                        & $12.7 \pm 3.2$ \\
JPLD + Bx/By/Bz \& vx/vy/vz (Omniweb)  & $15.5 \pm 12.6$ \\
JPLD + Orbital Mechanics + Quasi-Dipole & $\mathbf{9.2 \pm 4.5}$ \\
JPLD + All (Non-Residual Target)       & $18.8 \pm 10.7$ \\
JPLD + All                             & $10.7 \pm 4.5$ \\
\hline
\end{tabular}
\label{tab:ablation_inputs}
\end{table*}

\section{Summary}

The proposed IonCast GNN model accurately forecasts global TEC over a long lead time while processing heterogeneous multi-source data. To the best of our knowledge, IonCast GNN is the first machine learning approach to combine heterogeneous solar and geomagnetic driver data with TEC observations to forecast global TEC. Due to the increasing dependence on satellite infrastructure, an accurate forecast model for the ionosphere is necessary. IonCast GNN allows for the global forecast of TEC over long lead times with high accuracy, even during geomagnetic storm events. The preliminary results of our model show promise for the field of ionospheric forecasting and open the door for further exploration.


\section*{Acknowledgments}
This research is the result of the Frontier Development Lab, Heliolab a partnership between NASA, Trillium Technologies Inc. (USA), Google Cloud, NVIDIA and Pasteur Labs, Contract No. 80GSFC23CA040. A portion of research was carried out at the Jet Propulsion Laboratory, California Institute of Technology, under a contract with NASA. The authors thank Andrew Smith and Umaa D. Rebbapragada for their valuable insights, NASA's Goddard Space Flight Center, and NASA's Jet Propulsion Laboratory for their continuing support. 
 

\appendix


\pagebreak
\bibliographystyle{unsrt}
\bibliography{references.bib}

\begin{thebibliography}{10}

\bibitem{kelley2009}
Michael~C Kelley.
\newblock {\em The Earth's ionosphere: Plasma physics and electrodynamics}, volume~96.
\newblock Academic press, 2009.

\bibitem{Tsunoda1988}
Roland~T Tsunoda.
\newblock High-latitude f region irregularities: A review and synthesis.
\newblock {\em Reviews of Geophysics}, 26(4):719--760, 1988.

\bibitem{Kintner1976}
Jr. {Kintner}, P.~M.
\newblock {Observations of velocity shear driven plasma turbulence}.
\newblock {\em Journal of Geophysical Research}, 81(A28):5114--5122, October 1976.

\bibitem{Pulkkinen2017}
Antti Pulkkinen, E~Bernabeu, A~Thomson, A~Viljanen, R~Pirjola, D~Boteler, J~Eichner, PJ~Cilliers, D~Welling, NP~Savani, et~al.
\newblock Geomagnetically induced currents: Science, engineering, and applications readiness.
\newblock {\em Space weather}, 15(7):828--856, 2017.

\bibitem{Kataoka2022}
Ryuho Kataoka, Daikou Shiota, Hitoshi Fujiwara, Hidekatsu Jin, Chihiro Tao, Hiroyuki Shinagawa, and Yasunobu Miyoshi.
\newblock Unexpected space weather causing the reentry of 38 starlink satellites in february 2022.
\newblock {\em Journal of Space Weather and Space Climate}, 12:41, 2022.

\bibitem{Berger2020}
Thomas~E Berger, MJ~Holzinger, EK~Sutton, and JP~Thayer.
\newblock Flying through uncertainty.
\newblock {\em Space Weather}, 18(1):e2019SW002373, 2020.

\bibitem{Jakowski2011}
Norbert Jakowski, C~Mayer, MM~Hoque, and V~Wilken.
\newblock Total electron content models and their use in ionosphere monitoring.
\newblock {\em Radio Science}, 46(06):1--11, 2011.

\bibitem{Martire2024}
L{\'e}o Martire, Thomas~F Runge, Xing Meng, Siddharth Krishnamoorthy, Panagiotis Vergados, Anthony~J Mannucci, Olga~P Verkhoglyadova, Attila Komj{\'a}thy, Angelyn~W Moore, Robert~F Meyer, et~al.
\newblock The jpl-gim algorithm and products: multi-gnss high-rate global mapping of total electron content.
\newblock {\em Journal of Geodesy}, 98(5), 2024.

\bibitem{Bilitza2011}
Dieter Bilitza, Lee-Anne McKinnell, Bodo Reinisch, and Tim Fuller-Rowell.
\newblock The international reference ionosphere today and in the future.
\newblock {\em Journal of Geodesy}, 85(12):909--920, 2011.

\bibitem{Bilitza2017}
Dieter Bilitza, David Altadill, Vladimir Truhlik, Valentin Shubin, Ivan Galkin, Bodo Reinisch, and Xueqin Huang.
\newblock International reference ionosphere 2016: From ionospheric climate to real-time weather predictions.
\newblock {\em Space weather}, 15(2):418--429, 2017.

\bibitem{Ridley2006}
AJ~Ridley, Yue Deng, and G~T{\'o}th.
\newblock The global ionosphere--thermosphere model.
\newblock {\em Journal of Atmospheric and Solar-Terrestrial Physics}, 68(8):839--864, 2006.

\bibitem{Bilitza2022}
Dieter Bilitza, Michael Pezzopane, Vladimir Truhlik, David Altadill, Bodo~W Reinisch, and Alessio Pignalberi.
\newblock The international reference ionosphere model: A review and description of an ionospheric benchmark.
\newblock {\em Reviews of geophysics}, 60(4):e2022RG000792, 2022.

\bibitem{Zhukov2021}
Aleksei~V Zhukov, Yury~V Yasyukevich, and Aleksei~E Bykov.
\newblock Gimli: Global ionospheric total electron content model based on machine learning.
\newblock {\em GPS Solutions}, 25(1):19, 2021.

\bibitem{Smirnov2023}
A~Smirnov, Y~Shprits, F~Prol, H~L{\"u}hr, M~Berrendorf, I~Zhelavskaya, and C~Xiong.
\newblock A novel neural network model of earth’s topside ionosphere. sci rep 13: 1303, 2023.

\bibitem{Xiong2021}
Pan Xiong, Dulin Zhai, Cheng Long, Huiyu Zhou, Xuemin Zhang, and Xuhui Shen.
\newblock Long short-term memory neural network for ionospheric total electron content forecasting over china.
\newblock {\em Space Weather}, 19(4):e2020SW002706, 2021.

\bibitem{liu2020}
Lei Liu, Shasha Zou, Yibin Yao, and Zihan Wang.
\newblock Forecasting global ionospheric tec using deep learning approach.
\newblock {\em Space Weather}, 18(11):e2020SW002501, 2020.
\newblock e2020SW002501 2020SW002501.

\bibitem{ren2023}
Xiaodong Ren, Pengxin Yang, Dengkui Mei, Hang Liu, Guozhen Xu, and Yue Dong.
\newblock Global ionospheric tec forecasting for geomagnetic storm time using a deep learning-based multi-model ensemble method.
\newblock {\em Space Weather}, 21(3):e2022SW003231, 2023.
\newblock e2022SW003231 2022SW003231.

\bibitem{Mannucci1998}
AJ~Mannucci, BD~Wilson, DN~Yuan, CH~Ho, UJ~Lindqwister, and TF~Runge.
\newblock A global mapping technique for gps-derived ionospheric total electron content measurements.
\newblock {\em Radio science}, 33(3):565--582, 1998.

\bibitem{jpld}
Olga Verkhoglyadova and Xing Meng.
\newblock Global ionospheric maps for research -- jpld data product.
\newblock \url{https://sideshow.jpl.nasa.gov/pub/iono_daily/gim_for_research/jpld/}, April 2024.
\newblock Last updated: 8 Apr 2024. Government sponsorship acknowledged.

\bibitem{Stone1998}
Edward~C Stone, AM~Frandsen, RA~Mewaldt, ER~Christian, D~Margolies, JF~Ormes, and F~Snow.
\newblock The advanced composition explorer.
\newblock {\em Space Science Reviews}, 86(1):1--22, 1998.

\bibitem{Nishida1992}
A~Nishida, K~Uesugi, I~Nakatani, T~Mukai, DH~Fairfield, and MH~Acuna.
\newblock Geotail mission to explore earth's magnetotail.
\newblock {\em Eos, Transactions American Geophysical Union}, 73(40):425--429, 1992.

\bibitem{Wilson2021}
Lynn~B Wilson~III, Alexandra~L Brosius, Natchimuthuk Gopalswamy, Teresa Nieves-Chinchilla, Adam Szabo, Kevin Hurley, Tai Phan, Justin~C Kasper, No{\'e} Lugaz, Ian~G Richardson, et~al.
\newblock A quarter century of wind spacecraft discoveries, 2021.

\bibitem{Matzka21}
J.~{Matzka}, C.~{Stolle}, Y.~{Yamazaki}, O.~{Bronkalla}, and A.~{Morschhauser}.
\newblock {The Geomagnetic Kp Index and Derived Indices of Geomagnetic Activity}.
\newblock {\em Space Weather}, 19(5):e2020SW002641, May 2021.

\bibitem{Tobiska2012}
W~Kent Tobiska, BR~Bowman, and SD~Bouwer.
\newblock Solar and geomagnetic indices for thermospheric density models.
\newblock {\em COSPAR International Reference Atmosphere, edited by Rees D. and Tobiska WK}, 2012.

\bibitem{Laundal2017}
Karl~Magnus Laundal and Arthur~D Richmond.
\newblock Magnetic coordinate systems.
\newblock {\em Space science reviews}, 206(1):27--59, 2017.

\bibitem{Giorgini1996}
JD~Giorgini, DK~Yeomans, AB~Chamberlin, PW~Chodas, RA~Jacobson, MS~Keesey, JH~Lieske, SJ~Ostro, EM~Standish, and RN~Wimberly.
\newblock Jpl's on-line solar system data service.
\newblock In {\em AAS/Division for Planetary Sciences Meeting Abstracts\# 28}, volume~28, pages 25--04, 1996.

\bibitem{Hochreiter1997}
Sepp Hochreiter and Jürgen Schmidhuber.
\newblock Long short-term memory.
\newblock {\em Neural Computation}, 9(8):1735--1780, 11 1997.

\bibitem{Lam2023}
Remi Lam, Alvaro Sanchez-Gonzalez, Matthew Willson, Peter Wirnsberger, Meire Fortunato, Ferran Alet, Suman Ravuri, Timo Ewalds, Zach Eaton-Rosen, Weihua Hu, Alexander Merose, Stephan Hoyer, George Holland, Oriol Vinyals, Jacklynn Stott, Alexander Pritzel, Shakir Mohamed, and Peter Battaglia.
\newblock Learning skillful medium-range global weather forecasting.
\newblock {\em Science}, 382(6677):1416--1421, 2023.

\bibitem{wandb}
Lukas Biewald.
\newblock Experiment tracking with weights and biases, 2020.
\newblock Software available from wandb.com.

\bibitem{physicsnemo}
Physicsnemo | nvidia developer.
\newblock [Online; accessed 2025-08-30].

\end{thebibliography}

\newpage
\section*{Appendix}

\begin{table}[h!]
\centering
\caption{Comparison of IonCast LSTM and GNN model performance across growing forecast horizons (1, 6, and 12 hours) against IRI \cite{Bilitza2011}. Values represent the RMSE averaged over the entire 48-time step (12-hour) period and all latitudes/longitudes. Bold values are the lowest RMSE for each event level (G0, G2, and G4).}
\resizebox{\textwidth}{!}{%
\begin{tabular}{lcccccccc}
\toprule
\textbf{Event Level} &
\textbf{IRI} &
\multicolumn{2}{c}{\textbf{1 Hour}} &
\multicolumn{2}{c}{\textbf{6 Hours}} &
\multicolumn{2}{c}{\textbf{12 Hours}} \\
& & LSTM & GNN & LSTM & GNN & LSTM & GNN \\
\midrule
G0 - 2018-04-20 & $5.32$ & $4.39$ & $\textbf{1.34}$ & $8.94$ & $3.44$ & $8.26$ & $10.89$ \\
G2 - 2015-09-07 & $6.24$ & $6.01$ & $\textbf{2.36}$ & $11.82$ & $5.72$ & $11.50$ & $17.14$ \\
G4 - 2023-04-23  & $16.59$ & $14.28$ & $\textbf{5.82}$ & $27.06$ & $13.3$ & $25.72$ & $36.21$ \\
\bottomrule
\end{tabular}}
\label{tab:iri_comparison}
\end{table}

\begin{figure}[h!]
    \noindent
    \center
    \makebox[\textwidth][c]{\includegraphics[width=1.25\textwidth]{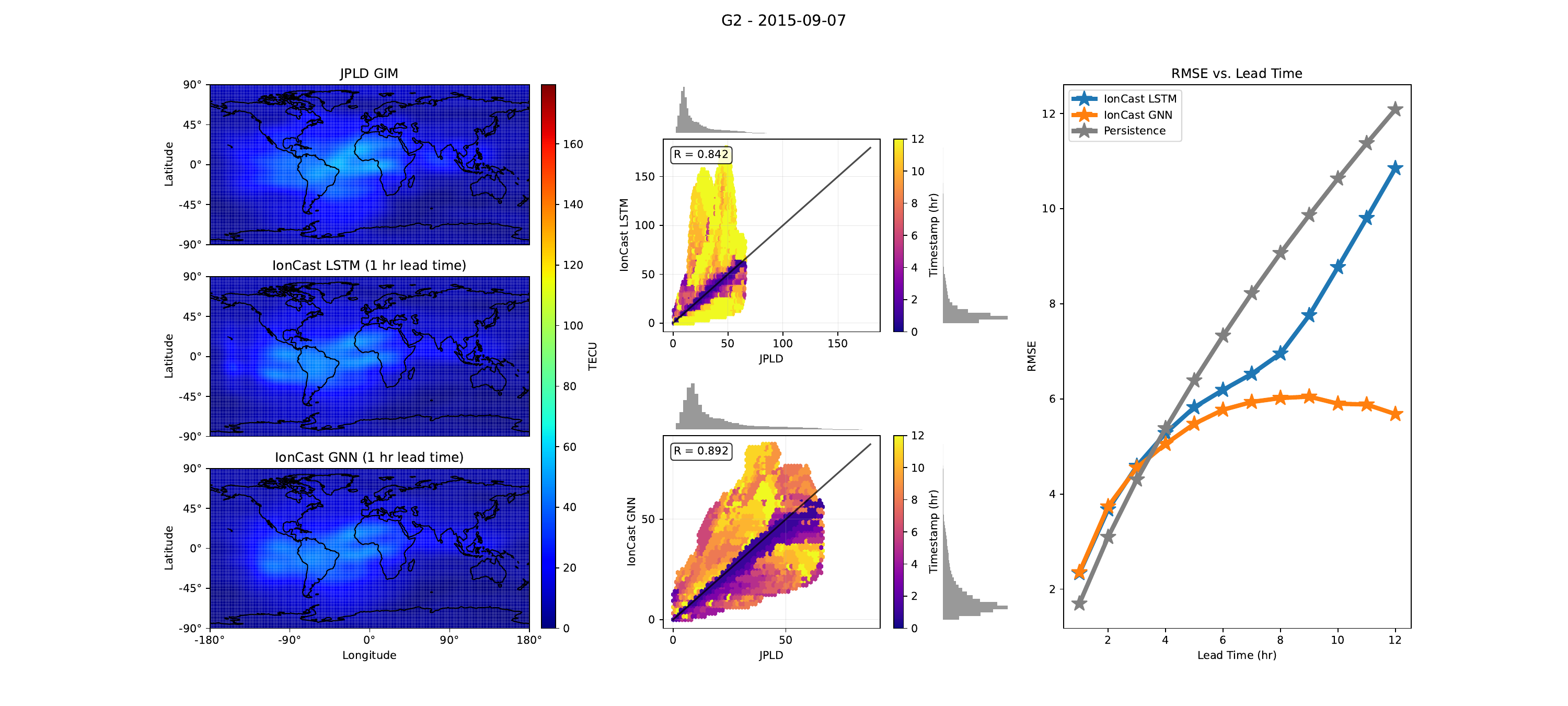}}
    \caption{Model predictions plotted and evaluated over various forecast lead times for a moderate geomagnetic storm (G2 event). The IonCast GNN model shown is the same as in the main body. The IonCast LSTM model is trained on a longer context window (16 15-minute cadence time steps) and a larger amount of training data (trains on every 32nd sequence of 4 hours, thus sequences skip 8 hours between start and end dates) than the 8-context and 256 date dilation model evaluated in the main text. \emph{Left:} Subplots show ground truth data from JPL (upper) plotted against predictions by the IonCast LSTM (middle) and IonCast GNN (lower) models at a 60-minute lead time. The colorbar has units TECU (TEC Units). \emph{Center:} The joint distribution between JPL (x-axis) and predictions (y-axis) over a 12-hour long-horizon forecast. The diagonal indicates perfect prediction; points below (above) the diagonal correspond to underprediction (overprediction). The colorbar encodes the mode of the forecast time (minutes) at which the prediction occurs, highlighting the error growth across the sequence. Marginal histograms show the distributions of ground truth and predicted values throughout the forecast. \emph{Right:} The plot compares the average global RMSE of the IonCast LSTM (blue), IonCast GNN (orange), and persistence model predictions across growing forecast horizons. }
    \label{fig:results-moderate-goodlstm}
\end{figure}

\pagebreak

\begin{figure}[h!]
    \noindent
    \center
    \makebox[\textwidth][c]{\includegraphics[width=1\textwidth]{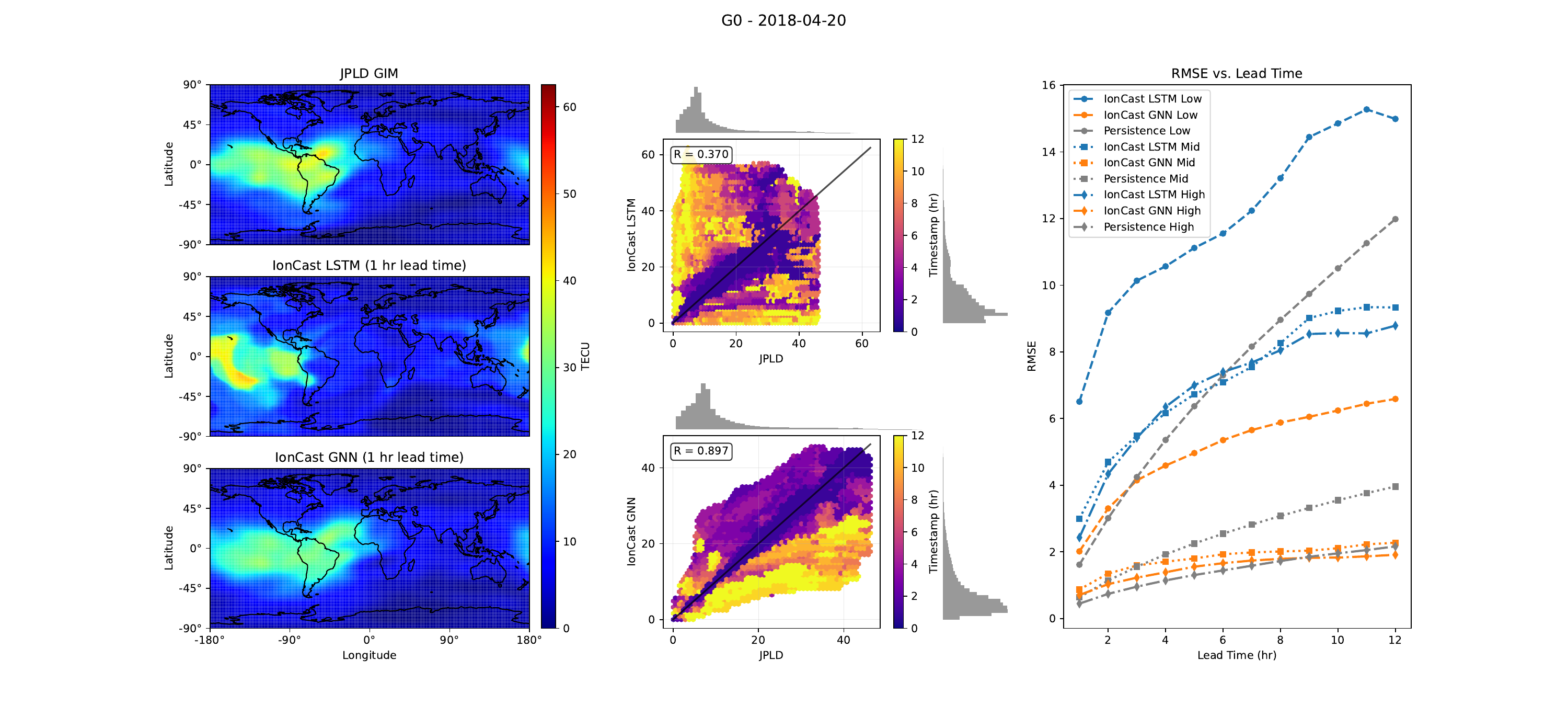}} 
    \makebox[\textwidth][c]{\includegraphics[width=1\textwidth]{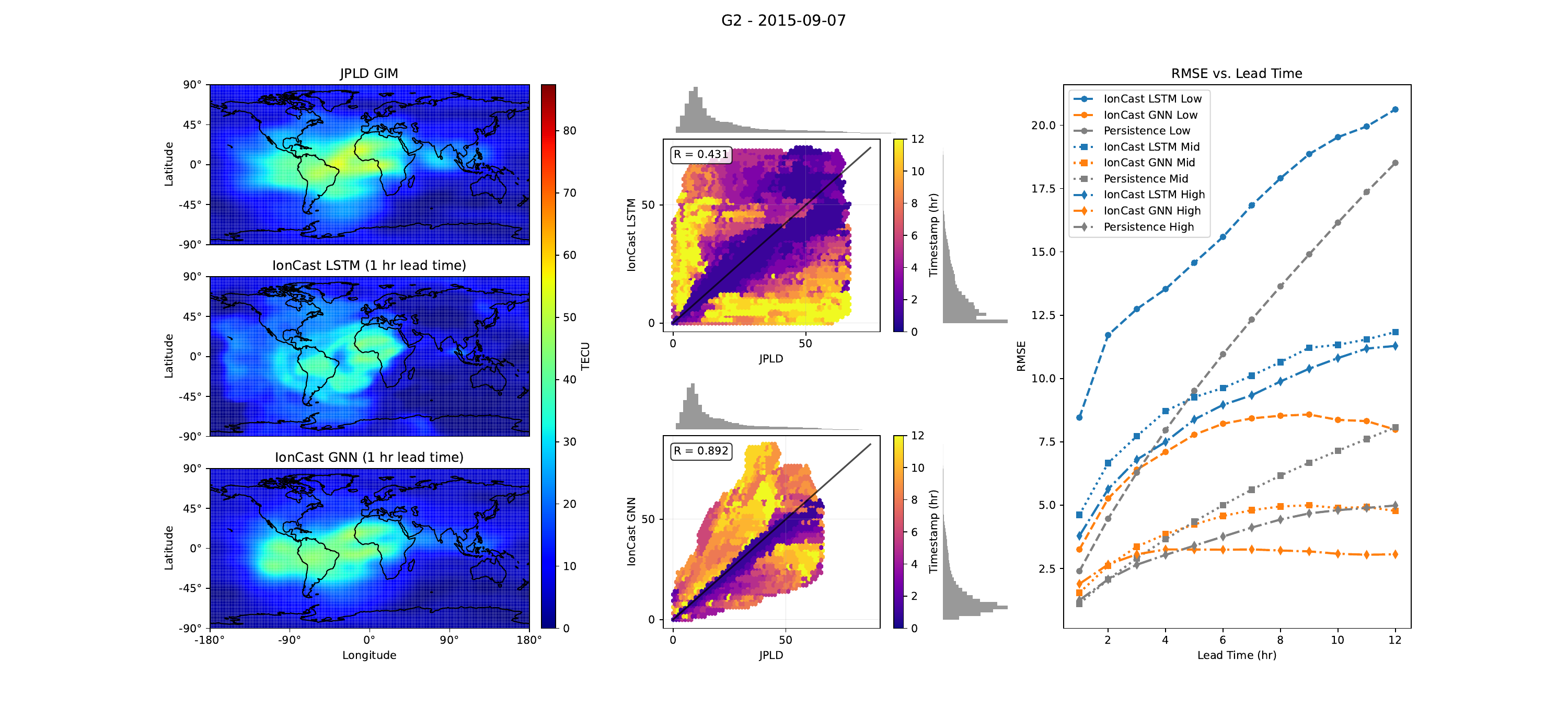}} 
    \makebox[\textwidth][c]{\includegraphics[width=1\textwidth]{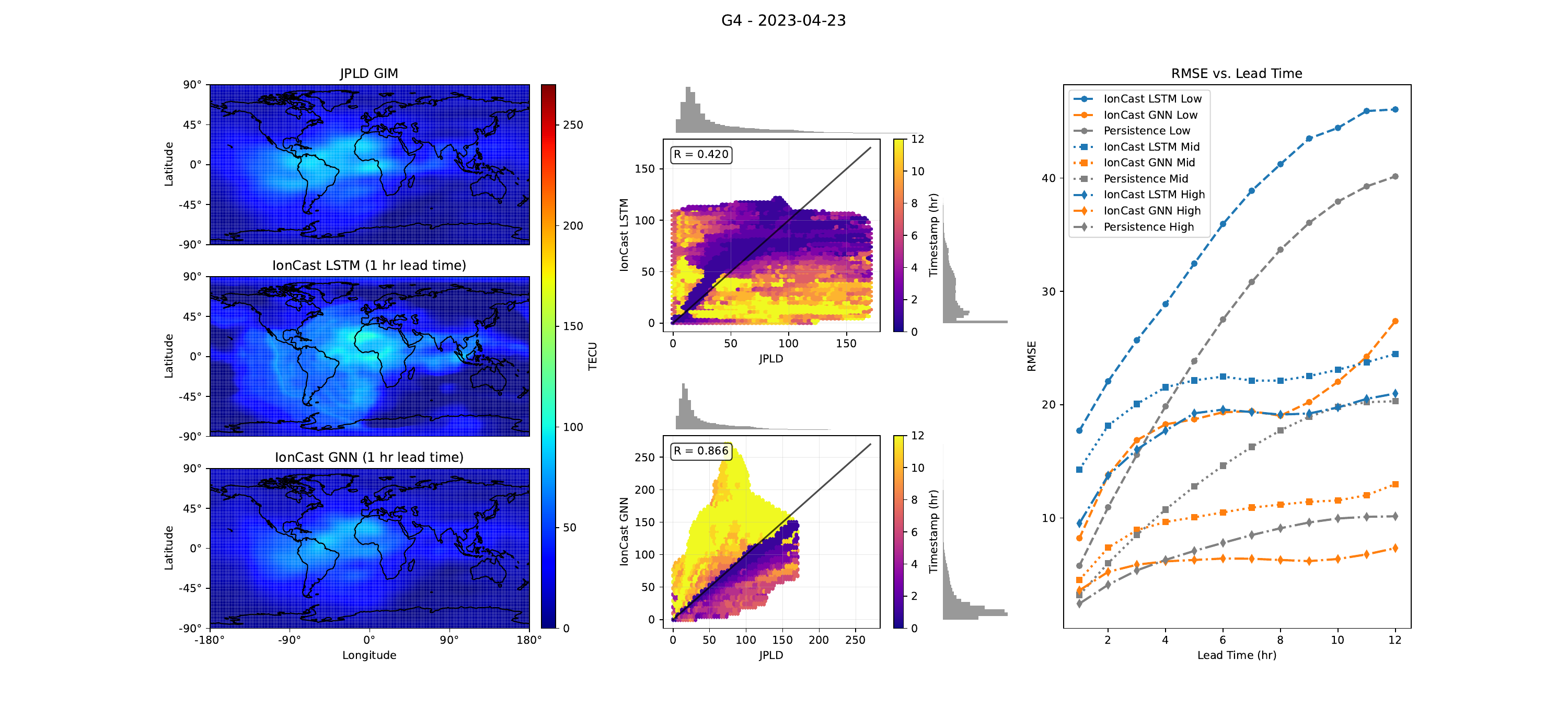}}
    \caption{Model predictions plotted and evaluated over various forecast lead times for a quiet ionospheric condition in the top figure (G0 event), a moderate (G2) and a severe geomagnetic event (G4 event) in the mid and bottom figure. \emph{Left} and \emph{Center} panels have the same models and meaning of the one shown in Figure 1. \emph{Right:} The plot on the right compares the RMSE of the IonCast LSTM (blue), IonCast GNN (orange), and persistence model predictions across growing forecast horizons, split across low ($30^{\circ}$ South to $30^{\circ}$ North), mid ($60^{\circ}$  to $30^{\circ}$ South and $30^{\circ}$ to $60^{\circ}$ North), and high latitudes ($60^{\circ}$ to $90^{\circ}$ South and $60^{\circ}$ to $90^{\circ}$ North).}
    \label{fig:results-quiet-moderate-severe}
\end{figure}


\begin{table}[h!]
\centering
\caption{Performance comparison for models trained during different solar cycle phases. There is only one G5 event present in the 2010-2024 date range. Bold values are the lowest RMSE for each event level.}
\resizebox{\textwidth}{!}{%
\begin{tabular}{lccccccc}
\toprule
\textbf{Data Range} & \makecell{\textbf{RMSE}\\\textbf{All Events}} & \makecell{\textbf{RMSE}\\\textbf{G0 Events}} &
\makecell{\textbf{RMSE}\\\textbf{G1 Events}} &
\makecell{\textbf{RMSE}\\\textbf{G2 Events}} &
\makecell{\textbf{RMSE}\\\textbf{G3 Events}} &
\makecell{\textbf{RMSE}\\\textbf{G4 Events}} &
\makecell{\textbf{RMSE}\\\textbf{G5 Events}} \\
\midrule

Solar Maximum 2013–2015 & $\textbf{8.63} \pm \textbf{4.34}$ & $\textbf{5.79} \pm \textbf{0.53}$ & $7.09 \pm 1.22$ & $\textbf{6.57} \pm \textbf{1.57}$ & $7.50 \pm 1.82$ & $\textbf{11.36} \pm \textbf{5.57}$ & $18.23$ \\
Solar Minimum 2018–2020 & $63.35 \pm 58.60$ & $7.31 \pm 3.76$ & $58.19 \pm 40.21$ & $49.61 \pm 42.55$ & $31.06 \pm 32.49$ & $148.42 \pm 40.12$ & $60.11$ \\
Full Dataset 2010–2024  & $8.87 \pm 4.44$ & $5.96 \pm 3.14$ & $\textbf{6.31} \pm \textbf{2.56}$ & $10.53 \pm 5.11$ & $\textbf{7.50} \pm \textbf{1.09}$ & $11.81 \pm 3.37$ & $\textbf{17.79}$ \\
\bottomrule
\end{tabular}}
\label{tab:solar_minmax}
\vspace{-4mm}
\end{table}

\begin{figure}[h!]
    \noindent
    \center
    \makebox[\textwidth][c]{\includegraphics[width=1\textwidth]{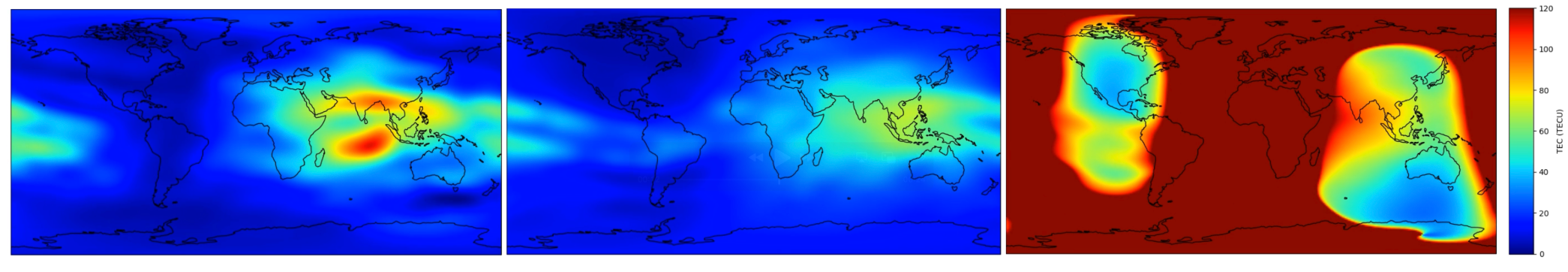}}    
    \caption{Solar cycle model predictions at a 360-minute lead time for a G4 level event. (Left) JPLD Ground Truth, (Middle) Model trained on solar maximum date range (2013-2015), (Right) Model trained on solar minimum date range (2018-2020).}
    \label{fig:solar-max-min}
\end{figure}

\begin{figure}[h!]
    \noindent
    \center
    \makebox[\textwidth][c]{\includegraphics[width=1\textwidth]{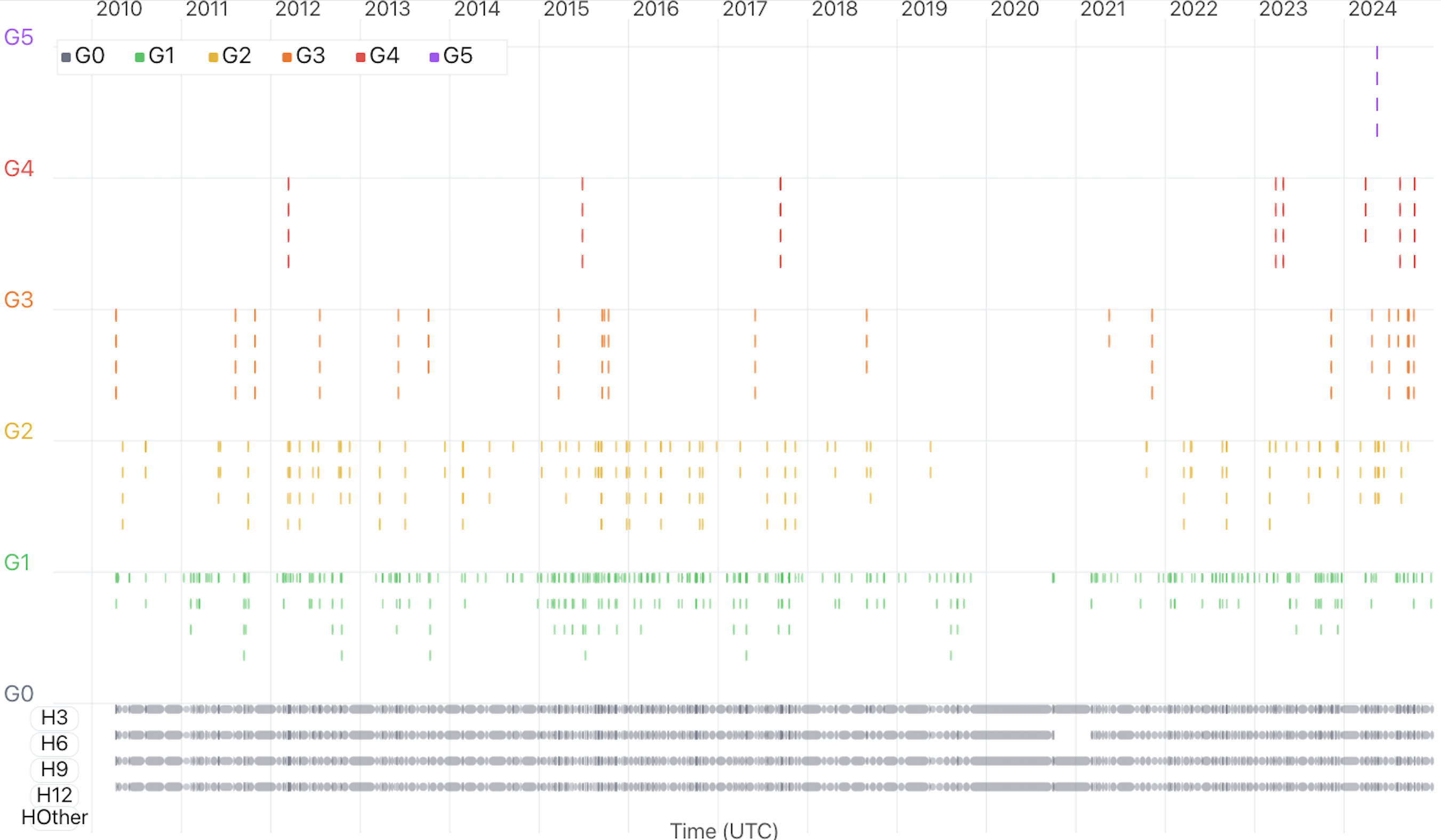}}    
    \caption{Distribution of G-level events over the data range (2010-2024). The x and y axes represent the time (years) and the intensity of the event (G-level), respectively. Each class bin in the y-axis is then divided into four segments, which correspond to the event duration, as shown in the lower part of the plot.}
    \label{fig:solar-event-catalog}
    \vspace{-4mm}
\end{figure}

The purpose of this ablation study was to examine how training on different portions of the solar cycle influences model performance and if models trained on local time periods can better predict specific event classes. All three model runs shown in Table~\ref{tab:solar_minmax}, were trained with the same set of hyperparameters, only varying in the date ranges seen in training. To ensure all three models received the same number of training samples, training sequences were more densely sampled for the models trained on the solar minimum and maximum date ranges compared to the model trained on the full date range. The solar maximum model performance is comparable to the model trained on the full date range. However, the solar minimum model struggles in all conditions other than G0 events. This seems to be caused by the lack of non-quiet solar events present in the solar minimum training range as seen in Figure~\ref{fig:solar-event-catalog}, leading to the poor generalization of the solar minimum model evaluated on events out of its training distribution Figure~\ref{fig:solar-max-min}.
\pagebreak

\begin{figure}
\centering
\begin{minipage}{.5\textwidth}
  \centering
  \includegraphics[width=1\linewidth]{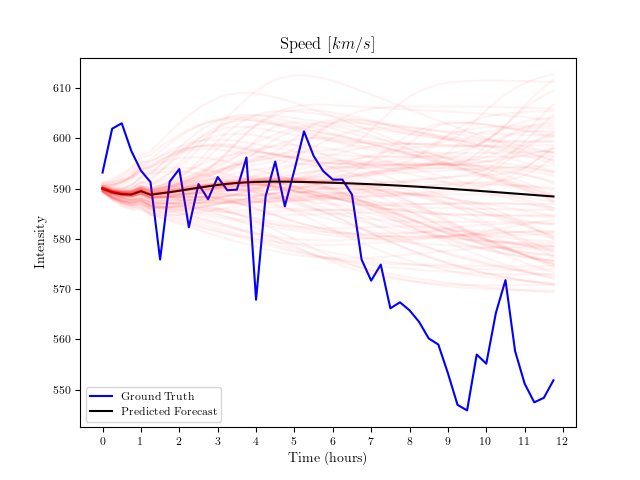}
  \label{fig:speed-forecast}
\end{minipage}%
\begin{minipage}{0.5\textwidth}
  \centering
  \includegraphics[width=1\linewidth]{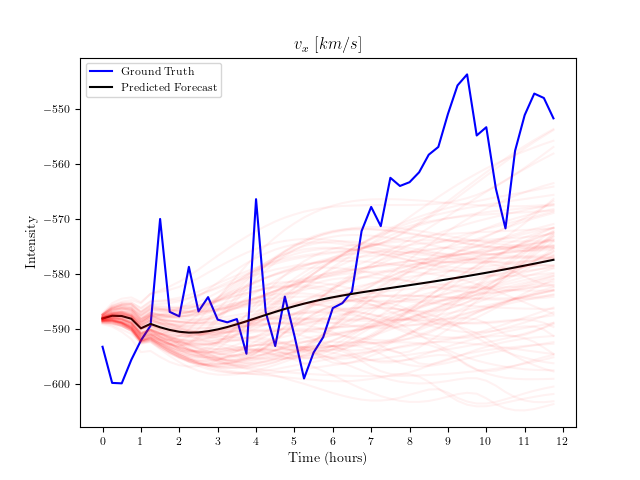}
  \label{fig:vx-forecast}
\end{minipage}
\caption{Selected forecasts corresponding to non-TEC 1-D drivers produced by the IonCast GNN model. Both forecasts correspond to the  2015-09-07 G2 event. The sampled forecasts (red) correspond to the forecasts generated at individual $1^{\circ} \times 1^{\circ}$ latitude-longitude nodes. The forecasts at 100 nodes were sampled to generate the red curves. The  predicted forecasts (black) correspond to the mean  forecast over all nodes of the latitude-longitude grid.}
\end{figure}

\begin{figure}[h!]
    \centering
    \makebox[\textwidth][c]{\includegraphics[width=\textwidth,trim=0 1cm 0 0,clip]{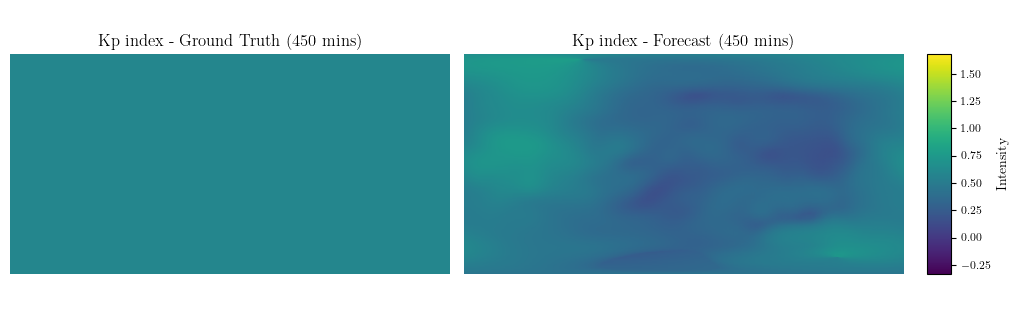}}
    \caption{Ground truth Kp index plotted against the global view of the pointwise ($1^{\circ}\times 1^{\circ}$) Kp predictions produced by the IonCast GNN model. The forecast corresponds to the quiet ionospheric conditions on 2018-04-20 (G0 event). }
    \label{fig:global-kp-frame}
\end{figure}

While examining the global view of the channels corresponding to 1-D drivers, certain patterns emerge that resemble TEC structure. This structure is shown in Figure \ref{fig:global-kp-frame}. This suggests the model may use the redundancy in the global feature channels as registers to inform TEC predictions.

\end{document}